# Semi-supervised Classification: Cluster and label approach using Particle Swarm Optimization

Shahira Shaaban Azab
Department of Computer Sciences
Institute of Statistical Studies and Research (ISSR)
Cairo, Egypt

Mohamed Farouk Abdel Hady
Department of Computer Sciences
Institute of Statistical Studies and Research (ISSR)
Cairo, Egypt

Hesham Ahmed Hefny
Department of Computer Sciences
Institute of Statistical Studies and Research (ISSR)
Cairo, Egypt

*Abstract*—supervised classification predicts classes of objects using the knowledge learned during the training phase. This process requires learning from labeled samples. However, the labeled samples usually limited. Annotation process is annoying, tedious, expensive, and requires human experts. Meanwhile, unlabeled data is available and almost free. Semi-supervised learning approaches make use of both labeled and unlabeled data. This paper introduces cluster and label approach using PSO for semi-supervised classification. PSO is competitive to traditional clustering algorithms. A new local best PSO is presented to cluster the unlabeled data. The available labeled data guides the learning process. The experiments are conducted using four state-of-the-art datasets from different domains. The results compared with Label Propagation a popular semi-supervised classifier and two state-of-the-art supervised classification models, namely k-nearest neighbors and decision trees. The experiments show the efficiency of the proposed model.

## I. Importance of unlabeled data

The unlabeled data is important in the classification of various real world application. In some applications, the unlabeled data have drastic effects on the performance of the classifiers. In other applications, the classification process may not be possible without unlabeled data.

Consider a two class problem as an example. We have only one feature $x$, we have two labeled data points: one positive labeled data represented as small circles and the other point is a negative data point depicted as a cross. The unlabeled data depicted as dots, see fig1.

There is only one labeled example for each class. The decision boundary should be in the middle between positive and negative data points. Thus, the decision boundary is the solid line as shown in fig1(a). With the help of unlabeled data points, more accurate decision boundary can be found as in fig1(b).

In speech analysis, labeling acoustic signal needs human annotators to listen to a conversation and label the phoneme. Annotating an hour of speech requires 400 hours annotation time [1]. Avery boring and time-consuming task.

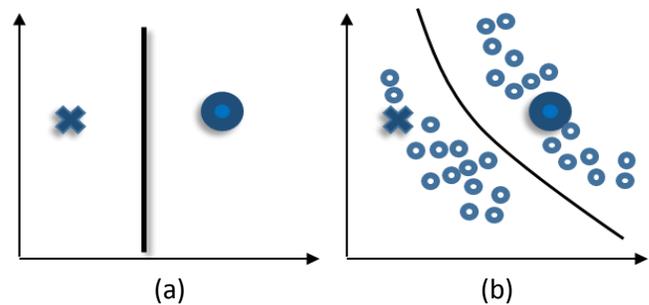

Fig. 1. Importance of Unlabeled Data

One application of semi-supervised classification is objects recognition in images. The semi-supervised classification is the most appropriate choice to address object recognition problems. Since we can use available data in the picture: shapes, textures, and colors to recognize objects of an image without any annotation[2]. Supervised classification requires a complicated process of labeling images. Thus, a small dataset usually less than 10,000 images is used. Unfortunately, this small dataset of labeled images is not representative of real world class distribution [2]. On the other hand, labeling a large dataset of images is a very expensive process.

In medical image analysis or computer-aided diagnosis (CAD), it is cheap and easy to have scanned images of the patients, but it is an expensive process to label them. Labeling process requires an expert such as a physician or radiologist to highlight the abnormal areas[3].

In some domains such as graphs, semi-supervised classification is the only possible classification methods because some nodes may have very few neighbors of a particular class[4].

In remote sensing application, high-resolution remote sensing sensors produce data in huge spectral bands. This data used to classify and understand Earth's materials. Since there are a large number of spectral bands and classes, a huge amount of labeled data is needed. However, this labeled data is expensive and requires experts[5].

In natural language, the parsing data instance is a sentence, and the class label is the corresponding parse tree. The training dataset of (sentence, parse tree) is known as a Treebank. Creating a Treebank is a very expensive and time-consuming process that required experts in linguistics. For creating 4000 sentences in the Penn Chinese Treebank, two years are needed[1]. Meanwhile, text can be easily found on World Wide Web.

The predicting structure of some protein sequence is a Semi-supervised problem. Identifying the structure of other proteins is expensive and require specialized apparatus to resolve. In fact, recognizing (3D) structure or of a single protein requires months and expert annotators while DNA sequences are available in the gene database[1], [6].

In documents categorization, we may not have a word "Obama" in the available small training set of the political category. However, using unlabeled documents that have "Obama" in them and they are classified as political documents because they have some shared features (words) with the training set. The new documents with the word "Obama" can be categorized in the political category with the help of the unlabeled data [4].

In handwritten digits recognition problem, there is a lot of variation for the same digit. Thus, the recognition process is very complicated. However, similarity can be found with the help of unlabeled data [1].

The rest of this paper is organized as follows: Section 2,3 presents a formal definition for supervised and semi-supervised classification respectively, section 4 introduces PSO for clustering and classification, section 5 explains proposed model, section 6 describes the experiments and results. Section 7 concludes this paper.

## II. SUPERVISED CLASSIFICATION

Supervised learning model is a function $f: \mathbb{X} \to \mathbb{y}$. $\{(x_i, y_i)\}_{i=1}^l$ Where $(x_i, y_i) \in \mathbb{X} * \mathbb{y}$ and $\{(x_i, y_i)\}_{i=1}^n$ is randomly and independently drawn from an unknown probability distribution $P_{\mathbb{X} \times \mathbb{y}}$. $\mathbb{X}$ represents features and $\mathbb{y}$ denotes class labels.

The objective of supervised learning is to find $f$ that can correctly predict the class $y$ from newly unseen $X$. Supervised learning model should choose $f \in F$ such that $f$ minimize prediction error. $\min_{i=1}^l (f(x_i) - y_i)$. The best $f$ is Bayes optimal classifier $\eta(x) = [\mathbb{E}(Y|X = x)]$. The performance improves with the increase in the training set size[7]. A limited number of labeled examples compromise the performance of the classifier.

## III. SEMI-SUPERVISED LEARNING (SSL)

SSL addresses the problem of rarely labeled data by using unlabeled data. SSL uses unlabeled data to identify the structure of the data. Semi-supervised learning is categorized as constrained clustering and semi-supervised classification(SSC).

Constrained Clustering: The main objective is to create better clusters by using both unlabeled data and make use of some of the supervised domain knowledge about this data e.g. link, must-links, cannot-links[1], [8].

Semi-supervised classification is a promising new direction in classification research [1], [6]. SSC uses a few labeled examples and massive unlabeled data. Labeling process for the training data especially when datasets are huge is a complicated, expensive, time-consuming process. As well, labeling process usually requires experts. SSC uses labeled data and make use of abundant unlabeled data to enhance classifiers performance.

Thus, SSC has both advantages of supervised and unsupervised learning by using both labels as well as the underlying structure of the data[2]. SSC is also known as classification with labeled and unlabeled data or partially labeled data classification [1].

The input of the SSC Model is:

1. Labeled training examples $\{(x_i, y_i)\}_{i=1}^l$. Where $(x_i, y_i) \in \mathbb{X} * \mathbb{y}$ are data points generated from the joint probability distribution $P_{\mathbb{X} \times \mathbb{y}}$, $X$ features Vector, and $y$ labels of the class.

2. Unlabeled training examples $\{x_i\}_{i=l+1}^{l+u}$. Where $X$ denotes features without corresponding labels generated from marginal distribution $P_{\mathbb{X}}$.

$l$ is the number of is labeled training examples, $u$ is the number of unlabeled examples. The number of unlabeled data is usually much more than labeled one $u \gg l$ because unlabeled data is cheap or even free. On the other side, labeled data is more expensive. The goal is to predict $p(c|x_l, x_u)$ and the evaluation metric used is the same metrics used for supervised classification i.e classification accuracy.

There are different models to solve the SSL problem such as:

Self-training: increase size of the training set by labeling unlabeled data. Then, the training set is enlarged by adding the most confident prediction to it.

Co-training: we have two classifiers. The features are represented into two disjoint subsets, and each of these subsets is sufficient to train a classifier. The most confident predictions of each classifier are added to the training set of the other classifier.

## IV. PARTICLE SWARM OPTIMIZATION(PSO)

### A. Global Particle Swarm Optimization(PSO)

PSO is a popular optimization algorithm. PSO mimics the foraging behavior of bird flocks. Swarm $W$ has $m$ particles $P_n$.

At time $t$, each particle has position $X_t$ and velocity $V_t$ for each dimension $d$. The position and velocity of each particles are updated using (1)-(3) where $x_{gbest}$ is the best position found by the swarm, $i$ Current particle index, $x_i, v_i$ are the position and velocity of the current particle respectively, $p_{best}$ represents the best position found by the particle, $g_{best}$ is the index of the global best particle in the entire swarm.

$$\boldsymbol{v_i(t+1) = \omega_i * v_i(t) + \varphi_1 * \left(p_{best_i}(t) - x_i(t)\right)} \quad (1)$$
$$\boldsymbol{+ \varphi_2 * (g_{best}(t) - x_i(t))}$$

$$x_i(t+1) = x_i(t) + v_i(t) \quad (2)$$

*B. PSO for Classification*

This section presents a brief survey about standard PSO for the classification problems. The numbers of research papers and results of experiments show the PSO is competitive to the standard classification algorithms[9]–[14].

PSO is an optimization algorithm which is very successful in handling continuous function optimization. PSO used primarily as a function optimization algorithm. Therefore, Classification problem is formulated as an optimization problem. Given a classification problem of a dataset with $C$ Classes and $D$ attributes, we can formulate it as a typical optimization problem of finding the optimal position of classes' centroids taking into consideration that each centroid has $D$ dimensions(features)[9].

Encoding: Each particle is represented as a vector of the centroids of classes in the dataset. Therefore, the global best particle is the proposed solution for the classification problem. There are $\eta$ particles in the swarm. Each particle $i$ in the swarm is represented by its velocity and position in different dimensions at time $t$.

Postion of particles is $(x_i^1, x_i^2, x_i^3, \ldots \ldots \ldots \ldots x_i^c)_t$.

velocity is encoded as $(v_i^1, v_i^2, v_i^3, \ldots \ldots \ldots \ldots v_i^c)_t$.

Where $x_i^j = \{x_{i1}^j, x_{i2}^j, x_{i3}^j, \ldots \ldots \ldots x_{id}^j\}$
$v_i^j = \{v_{i1}^j, v_{i2}^j, v_{i3}^j, \ldots \ldots \ldots v_{id}^j\}$ for J centroid and d features

In [9], Authors used three different fitness function for PSC as follows:

1. Percentage of misclassified instances of training set

$$\psi_1(i) = \frac{100.0}{m} \sum_{j=1}^{m} \delta(\vec{s}_k) \quad (3)$$

$$\delta(\vec{s}_k) \begin{cases} 1 & \text{incorrect class} \\ 0 & \text{correct class} \end{cases}$$

2. Sum of the distance of all the training set and the class.

$$\psi_2(i) = \frac{1}{m} \sum_{j=1}^{m} d(\vec{s}_k, \vec{x}_i) \quad (4)$$

3. Hybrid between equations **Error! Reference source not found.** and **Error! Reference source not found.**

$$\psi_3(i) = \frac{1}{2} \left( \frac{\psi_1(i)}{100.0} + \psi_2(i) \right) \quad (5)$$

is the time, $c$ represent class, k denotes k<sup>th</sup> example in the class $c$, $m$ is Size of training set (number of training points), $x_i^c$ is the potential **c**entroid of a class and Current position of particle $x$ for class c in d dimension, $\vec{s}_j$ example of the training set.

The experiment showed that $\psi_3$ is the best fitness function on 9 from 13 datasets[9]. The authors in [15] assure the superiority of $\psi_3$ in the experiments conducted in the context of content-based image retrieval using three fitness function. The results showed that $\psi_3$ is better in the convergence and the most suitable for this application. However, $\psi_2$ is the most common in the literature.

*C. PSO for Clustering*

PSC can be explained as follows[16]. Given a data set O with K clusters and D features. Each particle is encoded a vector of the centroids of clusters in the dataset, see fig 2. Hence, each particle is a potential solution to a clustering problem. Thus, the global best particle is the proposed solution for the clustering problem. Particles update their positions and velocities to obtain the optimal position for the centroids. Fitness function commonly used to evaluate the performance of particles is the minimum distance between points and potential centroids.

V. Semi-supervised PSO(SPSO)

This section presents the proposed SPSO. It uses LPSOC model of PSO for partition-based clustering[17]. LPSOC uses a pre-defined number of clusters $K$. Each neighborhood represents one of the clusters. The goal of the particles in each neighborhood is optimizing the position of the centroid of the cluster. LPSOC uses lbest model for PSO. This representation is simpler. Furthermore it not suffer from redundant representation of cluster solution, see fig 3. The information from available labeled data guides the particles of LPSOC in identifying clusters of objects in the datasets. Hence, the fitness function must satisfy two condition. First, the formed clusters must follow the information of the labeled data. Second, cluster should have the best silhouette score. Silhouette score is one of the common relative validity indices, see (6).

$$sil(i, K) = \left\{ \frac{b(i) - a(i)}{\max\{a(i), b(i)\}} \right\}_K \quad (6)$$

where object $i$ and $a(i)$ average similarity between data points in the cluster of object $i$ (the cohesion) and the average dissimilarity with other objects in other clusters is $b(i)$ (the separation).

The detailed algorithm of the proposed model is as follows:
Initialize the positions and the velocities for each particle in the swarm randomly.
Do while the termination condition is met
   For each neighborhood
      For each particle in the neighborhood
         For each data point
            Calculate distance (training data,particle)
            Assign data points to the nearest centroid
            Calculate the fitness of particles
            If disagree with the labeled data
               Assign random position and velocity
               Assign infinity fitness of the particle
   IF Current position better than best position
   THEN pbest position= current position.
   IF Current position is better than the global best
   THEN gbest index= current particle index.
   Update particle velocity using (1)
   Move the particle to a new position using (2).
For each unlabeled data points in impure clusters

Assign data points to the nearest neighbors

| $x_{i1}^1$ | $x_{i2}^1$ | ... | $x_{id}^1$ | $x_{i1}^2$ | $x_{i2}^2$ | ... | $x_{id}^2$ | ... | $x_{i1}^c$ | $x_{i2}^c$ | ... | $x_{i2}^c$ |
|---|---|---|---|---|---|---|---|---|---|---|---|---|
| centroid$_1$ ||| | centroid$_2$ |||| ... | centroid$_c$ ||||
| Particle $i$ ||||||||||||| |

Fig. 2. Representation of a particle of PSC

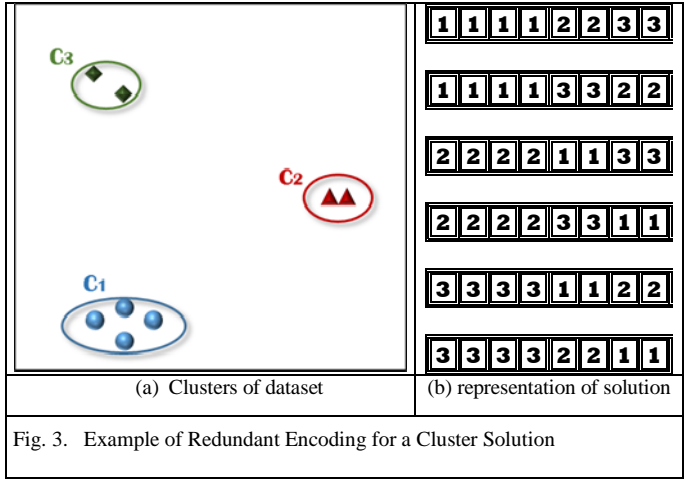

(a) Clusters of dataset    (b) representation of solution

Fig. 3. Example of Redundant Encoding for a Cluster Solution

## VI. EXPERIMENTS & RESULTS

The experiments is conducted using 4 state-of-the-art datasets available at[18]. The datasets used in the experiments are from different domains. Table I, and fig 4 illustrated the properties of the datasets. The proposed model is compared with two supervised learning models namely decision trees and k-nearest neighborhood and a semi-supervised learning model called label propagation.

The results of the average of 30 runs with different percentage of labels (1-10%, 11-20%, 21-40%, 41-90%) are listed in tables II-VI and the ROC curve are illustrated fig5-8 The results states that the proposed algorithm is superior to the decision tree, k-nearest neighborhood, and label propagation in almost all of the dataset. The superiority of the proposed algorithm are increasing with a low percentage of labels.

TABLE I.    PROPERTIES OF DATASETS

| Datasets | *Features* | *Samples* | *Classes* |
|---|---|---|---|
| haberman | 3 | 306 | 2 |
| Titanic | 3 | 2201 | 2 |
| Pima | 8 | 768 | 2 |
| Wisconsin | 10 | 699 | 2 |

TABLE II.    SIMULATION RESULTS FOR DATASETS 1-10% LABELS

| Datasets | F1_score ||||
|---|---|---|---|---|
| | *Tree* | *KNN* | *LP* | *SPSO* |
| haberman | 0.81 | 0.82 | 0.83 | **0.86** |
| titanic | 0.53 | 0.48 | 0.56 | **0.58** |
| pima | 0.74 | 0.80 | 0.78 | **0.89** |
| wisconsin | 0.84 | 0.79 | 0.87 | **0.98** |

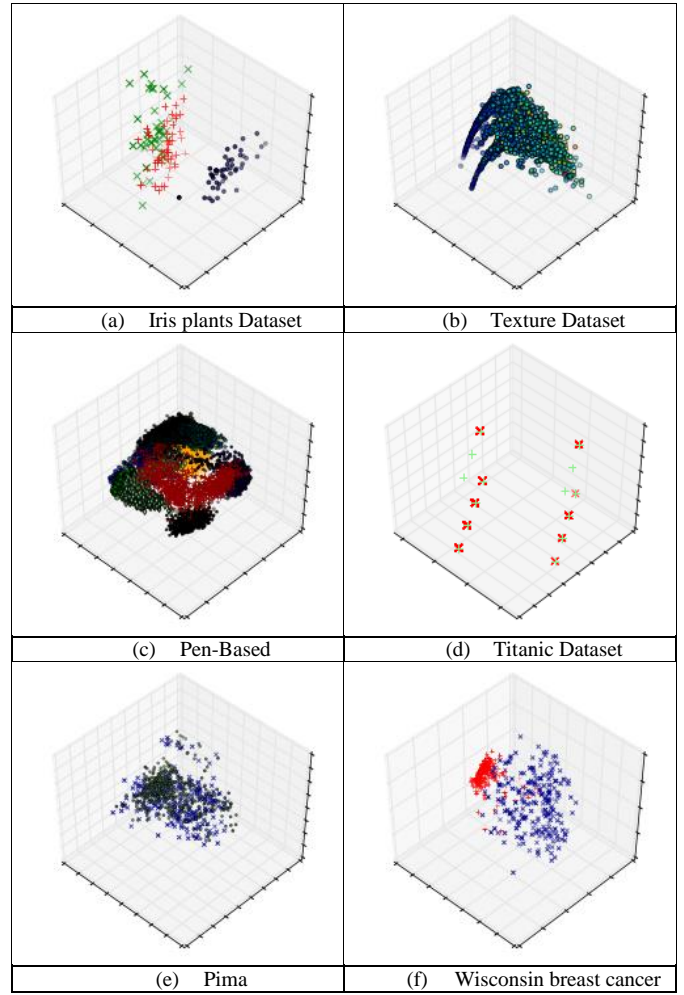

(a) Iris plants Dataset    (b) Texture Dataset
(c) Pen-Based    (d) Titanic Dataset
(e) Pima    (f) Wisconsin breast cancer

Fig. 4. Properties of datatsets

TABLE III.    SIMULATION RESULTS FOR DATASETS 11-20% LABELS

| Datasets | F1_score ||||
|---|---|---|---|---|
| | *Tree* | *KNN* | *LP* | *SPSO* |
| Haberman | 0.81 | 0.82 | 0.83 | **0.88** |
| Titanic | 0.58 | 0.55 | 0.54 | **0.59** |
| Pima | 0.79 | 0.77 | 0.80 | **0.80** |
| wisconsin | 0.95 | 0.95 | **0.97** | **0.97** |

TABLE IV.    SIMULATION RESULTS FOR DATASETS 21-40% LABELS

| Datasets | F1_score ||||
|---|---|---|---|---|
| | *Tree* | *KNN* | *LP* | *SPSO* |
| haberman | 0.73 | 0.80 | 0.82 | **0.84** |
| titanic | 0.56 | **0.57** | 0.54 | 0.55 |
| pima | 0.75 | 0.78 | 0.81 | **0.83** |
| wisconsin | 0.94 | **0.98** | 0.97 | **0.98** |

TABLE V.    SIMULATION RESULTS FOR DATASETS 41-90% LABELS

| Datasets | F1_score ||||
|---|---|---|---|---|
| | *Tree* | *KNN* | *LP* | *SPSO* |
| haberman | **0.86** | 0.80 | 0.77 | 0.73 |
| titanic | 0.56 | **0.77** | 0.52 | 0.53 |
| pima | 0.78 | **0.84** | 0.82 | 0.83 |
| wisconsin | 0.95 | **0.99** | 0.98 | 0.97 |

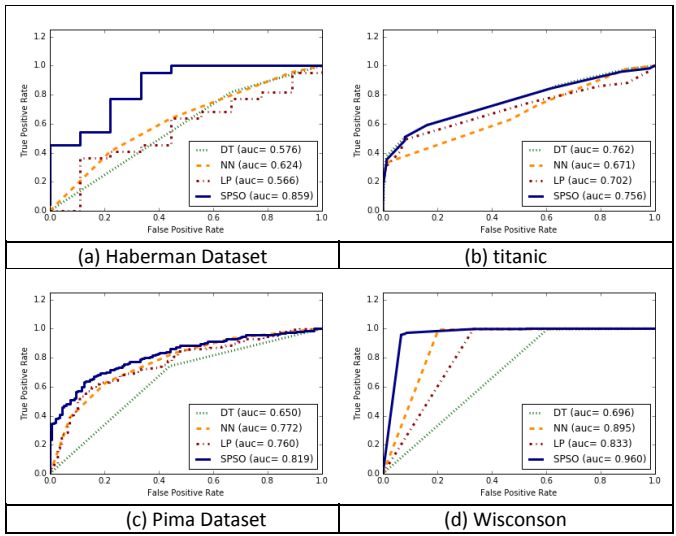

Fig. 5. ROC Curve of DataSets 1-10% Labels

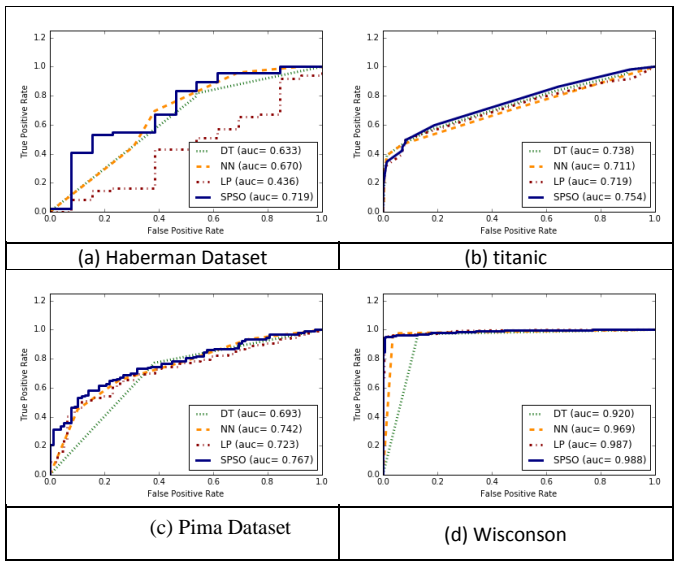

Fig. 6. ROC Curve of DataSets 11-20% Labels

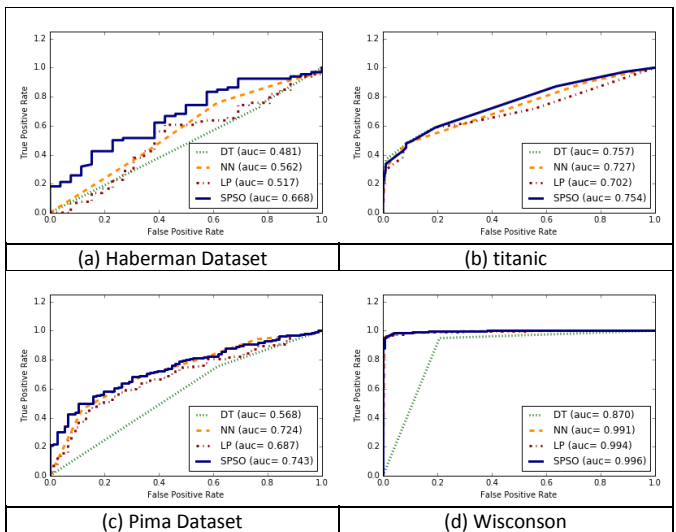

Fig. 7. ROC Curve of DataSets 21-40% Labels

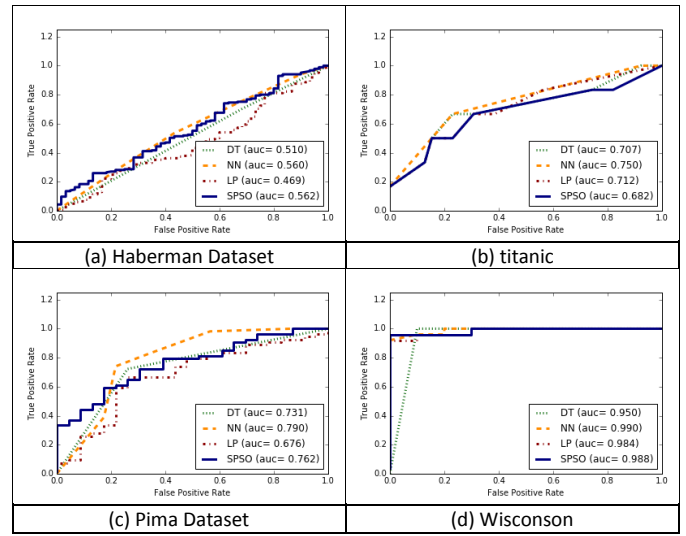

Fig. 8. ROC Curve of DataSets 41-90% Labels

## CONCLUSION

This paper presents an efficient semi-supervised model. The proposed model adopts the cluster-and-label model. Cluster-and-label model is one of the common mixture models. It tries to discover the distribution of unlabeled data and use the labeled data to classify the detected clusters. The proposed model uses the LPSOC to cluster data. Meanwhile, the LPOSC used the guidance provided by the labeled data to cluster unlabeled data accurately. With the help of both labeled and the cluster unlabeled data, the proposed semi-supervised model can classify existed unlabeled data (transductive setting) or predict the class of any unseen data (inductive setting). The experiments traditional datasets. Datasets are chosen from various domains. The effect of the ratio of available labeled data on the performance of proposed classifier is analyzed; the percentages of labeled data used are 1-10%, 11-20%, 21-40%, and 41-91%. The results are compared with two popular supervised algorithms, and one semi-supervised algorithm. The results assure the superiority of the proposed model especially with a limited amount of labeled data.